\def\year{2020}\relax
\pdfoutput=1
\documentclass[letterpaper]{article} 
\usepackage{aaai20}  
\usepackage{times}  
\usepackage{helvet} 
\usepackage{courier}  
\usepackage[hyphens]{url}  
\usepackage{graphicx} 
\usepackage{graphicx}  
\usepackage{amsmath}

\usepackage{graphics}
\usepackage{subfig}
\usepackage{float}
\usepackage{adjustbox}
\usepackage{nicefrac}       
\usepackage{microtype}      
\usepackage{amssymb }
\usepackage{algorithm}
\usepackage{algorithmic}

\title{Optimizing Discrete Spaces via Expensive Evaluations: \\ A Learning to Search Framework}
\author{Aryan Deshwal\textsuperscript{\rm 1}, Syrine Belakaria\textsuperscript{\rm 1}, Janardhan Rao Doppa\textsuperscript{\rm 1}, Alan Fern\textsuperscript{\rm 2}\\ 
\textsuperscript{\rm 1}School of EECS, Washington State University\\\textsuperscript{\rm 2}School of EECS, Oregon State University\\ 
{\{aryan.deshwal, syrine.belakaria, jana.doppa\}@wsu.edu, alan.fern@oregonstate.edu}
}
\begin{document}

\maketitle

\begin{abstract}
We consider the problem of optimizing expensive black-box functions over discrete spaces (e.g., sets, sequences, graphs). The key challenge is to select a sequence of combinatorial structures to evaluate, in order to identify high-performing structures as quickly as possible. Our main contribution is to introduce and evaluate a new learning-to-search framework for this problem called L2S-DISCO. The key insight is to employ search procedures guided by control knowledge at each step to select the next structure and to improve the control knowledge as new function evaluations are observed. We provide a concrete instantiation of L2S-DISCO for local search procedure and empirically evaluate it on diverse real-world benchmarks. Results show the efficacy of L2S-DISCO over state-of-the-art algorithms in solving complex optimization problems.
\end{abstract}

\section{Introduction}

Many scientific and engineering applications involve optimizing discrete spaces (e.g., sets, sequences, graphs) guided by expensive black-box function evaluations. For example, in the application of finding alloys with high creep-resistance, we need to search over subsets of a given set of candidate metals guided by physical lab experiments. Similarly, for 
designing application-specific integrated circuits, we need to search over candidate placements of processing elements and communication links to optimize performance as measured by expensive computational simulations. 

There is very limited work on optimizing discrete spaces via expensive evaluations as discussed in Section \ref{sec:related-work}. A popular and effective framework for optimizing expensive functions is Bayesian optimization (BO) \cite{BO-Survey:2016}.  
The key idea behind BO is to estimate a cheap surrogate model, e.g., a Gaussian Process \cite{GP-Book}, based on observed outcomes, which can be used as guidance for intelligently selecting the next evaluation points. Despite the huge successes of BO \cite{BO:NIPS2012,AutoWeka}, current approaches focus primarily on continuous optimization spaces and there is little principled work on discrete spaces. The first challenge in moving from continuous spaces to discrete spaces is to define an effective surrogate model over combinatorial structures. The second challenge is, given such a surrogate model, to search through the combinatorial space to identify the most promising next structure to evaluate.  Prior methods either employ relatively simple surrogate models that admit tractable optimization solvers for this search or use complex models with highly-heuristic search methods. Ideally, we would like an approach that can work with more complex models while following a more principled and effective search approach. 

In this paper, we introduce a new {\em learning-to-search (L2S) framework}, called L2S-DISCO, for selecting the sequence of combinatorial structures to evaluate during optimization. L2S-DISCO employs a combinatorial search procedure (e.g., local search with multiple restarts) guided by search control knowledge (e.g., heuristic function to select good starting states), and {\em continuously} improves the control knowledge using machine learning. Our approach can be viewed as online learning for hyper-heuristic search \cite{hyper-heuristics}. From this perspective, a primary contribution of this paper is the first application of hyper-heuristic search to BO. 

Our search-based perspective allows us to directly tune the search via learning during the optimization process and has several potential advantages: 1) High flexibility in defining search spaces over structures; 2) Easily handles domain constraints to search over ``valid'' structures; 3) Allows the incorporation of prior knowledge; and 4) Puts forth a new family of BO-style approaches with many future instantiations to explore in future work. We provide a concrete instantiation of L2S-DISCO for local search based optimization by specifying the form of training data, and a rank learning formulation to update the search heuristic for selecting promising starting states. Experimental results on diverse benchmarks show the efficacy of L2S-DISCO on complex real-world problems.

\vspace{1.0ex}

\noindent {\bf Contributions.} The key contributions of this paper include:
\begin{itemize}
\setlength\itemsep{0em}    
    \item A novel learning-to-search framework, L2S-DISCO, for optimizing expensive functions over discrete spaces, which integrates combinatorial-search with machine learning.
    \item A concrete instantiation of L2S-DISCO for local-search--based optimization.
    \item An evaluation of L2S-DISCO over diverse real-world benchmarks, showing advantages over the  state-of-the-art.  
\end{itemize}

\section{Problem Setup and Challenges}

\vspace{1.0ex}

\noindent {\bf Combinatorial space of structures.} Let $\mathcal{X}$ be a combinatorial space of objects to be optimized over, where each element $x \in \mathcal{X}$ is a discrete structure (e.g., set, sequence, graph). Without loss of generality, let each candidate structure $x \in \mathcal{X}$ be represented using $d$ discrete variables $v_1, v_2,\cdots,v_d$, where each variable $v_i$ take candidate values from a set $C(v_i)$. 
If each discrete variable takes $k$ candidate values, the size of the combinatorial space is $\mathcal{O}(k^d)$.

\vspace{1.0ex}

\noindent {\bf Problem definition.} We are given a combinatorial space of structures $\mathcal{X}$. We assume an unknown real-valued objective function $\mathcal{F}: \mathcal{X} \mapsto \Re$, which can evaluate each candidate structure $x \in \mathcal{X}$. 
For example, in alloys optimization application, $x$ is a set corresponding to material design and $\mathcal{F}(x)$ corresponds to running a physical lab experiment using additive manufacturing techniques. Conducting an experiment produces an evaluation $y$ = $\mathcal{F}(x)$ and is expensive in terms of the consumed resources. The main goal is to find a structure $x \in \mathcal{X}$ that approximately optimizes $\mathcal{F}$ by conducting a limited number of evaluations and observing their outcomes.

\vspace{1.0ex}

\noindent {\bf Bayesian optimization formulation and challenges.} Bayesian optimization (BO) methods \cite{BO-Survey:2016} build a surrogate {\em statistical model} $\mathcal{M}$, e.g., Gaussian Process (GP), from the training data of past function evaluations and employ it to sequentially select a sequence of inputs for evaluation to solve the problem (see Algorithm~\ref{alg:BO}). The selection of inputs is performed by optimizing an {\em acquisition function} $\mathcal{AF}$ that is parameterized by the current model $\mathcal{M}$ and input $x \in \mathcal{X}$ to score the utility of candidate inputs for evaluation. Some example acquisition functions include expected improvement (EI) \cite{EI:1998} and upper-confidence bound (UCB) \cite{UCB:ICML2010}. BO methods are mostly studied for continuous spaces $\mathcal{X} \subset \Re^d$. 
There is very limited work on BO methods to optimize discrete spaces (as discussed in the related work section). There are two key challenges in using BO for discrete spaces.

\begin{enumerate}
\setlength\itemsep{0em}  
    \item {\em Surrogate statistical modeling.} GPs are the popular choice for building statistical models in BO over continuous spaces. To handle discrete structures, we need an appropriate kernel that can compute the similarity between any pair of candidate structures $x, x' \in \mathcal{X}$. 
    One choice is to leverage the general recursive convolution framework \cite{recursive-convolution:TR-1999}. 
    The key idea is to recursively decompose the structured object into {\em atomic} sub-structures and define valid local kernels between them. For example, random walk kernels over graphs defined in terms of paths \cite{graph-kernels:JMLR2010}. Random forest (RF) models can be used as an alternate generic choice to handle discrete spaces. In this work, we employ RF models as part of our experiments.
    
    \item {\em Acquisition function optimization.} In each iteration of BO, we need to solve the following optimization problem to select the next candidate structure for evaluation.
    \begin{align}
    x_{next} = \mbox{arg}\,\mbox{max}_{x \in \mathcal{X}} \, \mathcal{AF}(\mathcal{M},x)
    \label{eq:afo}
    \end{align}
   The key challenge for discrete spaces is that, Equation~\ref{eq:afo} corresponds to solving a general combinatorial optimization problem. The effectiveness of BO critically depends on the accuracy of solving this optimization problem. In this paper, our main focus is on addressing this challenge (line 4 in Algorithm~\ref{alg:BO} given below) using a novel learning to search framework.
\end{enumerate}

\begin{algorithm}[H]
\caption{Bayesian Optimization framework}
\footnotesize
\textbf{Input}: $\mathcal{X}$ = Discrete space, 
$\mathcal{F}(x)$ = expensive objective function \\
\textbf{Output}: $\hat{x}_{best}$, the best uncovered structure from $\mathcal{X}$
\label{alg:BO}
\begin{algorithmic}[1]
\REPEAT
\STATE Learn the model: $\mathcal{M}_t \leftarrow$ \textsc{Learn}($\mathcal{D}_t$)
\STATE Compute the next structure to evaluate via acquisition function optimization: $x_{t+1} \leftarrow \mbox{arg}\,max_{x \in \mathcal{X}} \, \mathcal{AF}(\mathcal{M}_t,x)$
\STATE Evaluate objective function $\mathcal{F}(x)$ at $x_{t+1}$ to get $y_{t+1}$
\STATE Aggregate the data: $\mathcal{D}_{t+1} \leftarrow \mathcal{D}_{t} \cup \left\{(x_{t+1}, y_{t+1})\right\}$
\STATE $t \leftarrow t+1$
\UNTIL{convergence or maximum iterations}
\STATE $\hat{x}_{best} \leftarrow \mbox{arg}\,max_{x_t \in \mathcal{D}} \, y_t$
\STATE \textbf{return} the best uncovered structure $\hat{x}_{best}$
\end{algorithmic}
\end{algorithm}

\section{Related Work}
\label{sec:related-work}
\begin{figure*}[t!]
\centering
\subfloat[Subfigure 1 list of figures text][Contamination domain with UCB acquisition function.]{
\includegraphics[width=0.4\textwidth]{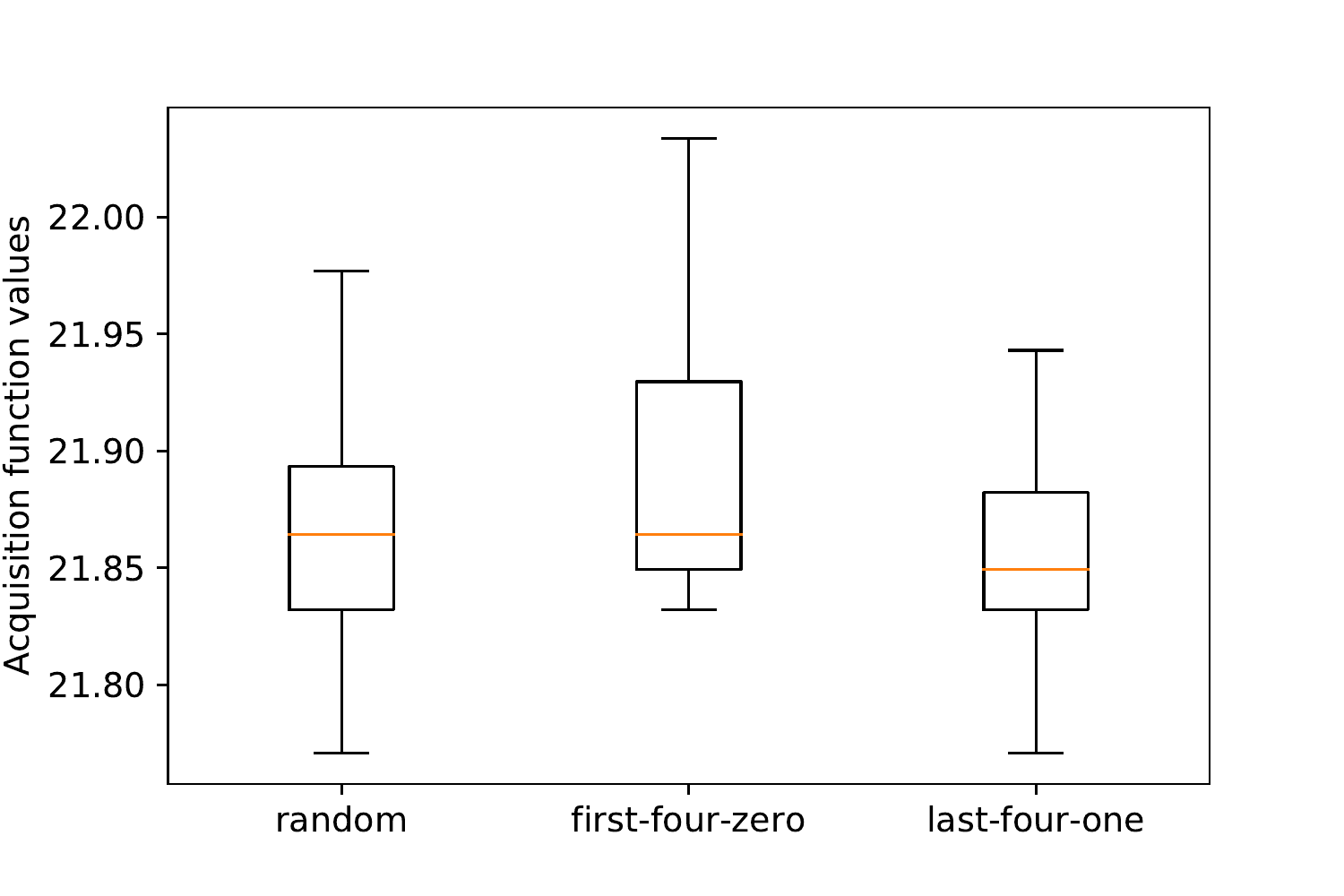}
\label{fig:contamination_empirical}}
\qquad
\subfloat[Subfigure 2 list of figures text][Ising domain with EI acquisition function.]{
\includegraphics[width=0.4\textwidth]{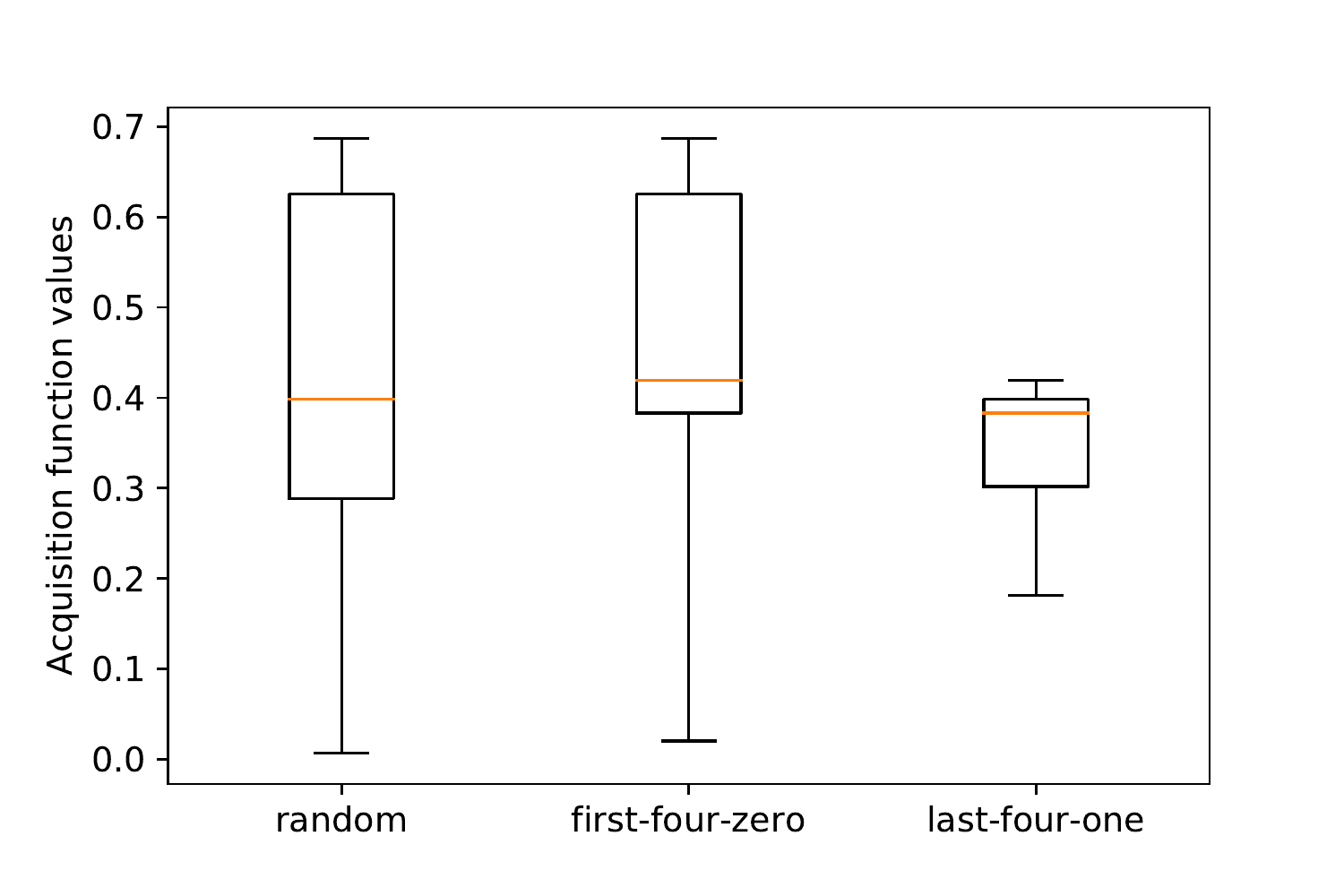}
\label{fig:ising_empirical}}
\caption{Empirical evidence to show how learning can be useful to solve acquisition function optimization. Boxplot shows final acquisition function values resulting from 100 runs of local search based optimization with three different restart strategies.} 
\label{fig:empirical_evidence}
\end{figure*}
There is very limited work on BO over discrete spaces. SMAC \cite{SMAC,SMAC:TR2010} is one canonical baseline which employs random forest as surrogate model and a {\em hand-designed} local search procedure for optimizing the acquisition function. BOCS \cite{BOCS} is a state-of-the-art method that employs a linear Bayesian model defined over binary variables as the surrogate model. The model is described as:
\begin{align}
    f_\alpha (x \in \mathcal{X}) = \alpha_0 + \sum_j \alpha_j v_j + \sum_{i, j > i} \alpha_{ij} v_i v_j
\end{align}
where $\mathcal{X} = \{0, 1\}^d$. The $\alpha$ variables drawn from a sparse prior, quantify the uncertainty  of the model. This linear model formulation over binary variables along with the usage of Thompson sampling as the acquisition function allows the acquisition function optimization in BOCS to be amenable to a semi-definite programming (SDP) solution. However, BOCS has several drawbacks. First, the simple model with second-order interactions may not suffice for optimization problems with complex interactions. Additionally, the model is very specific to binary variables. An extension to general categorical variables via one-hot encoding was provided in the supplementary section \cite{BOCS}, but this results in significant growth of input dimensions. Second, the SDP based acquisition function optimization solution is very specific to the case of binary variables and Thompson sampling, which severely limits its applicability. Additionally, one-hot encoding to handle categorical variables leads to poor scalability and loss of accuracy for BOCS. Third, the SDP based solver cannot handle complex constraints to select only valid structures. Indeed, we observe these shortcomings of BOCS in our experiments. In contrast, our proposed method can work with any choice of statistical model and acquisition function, and uses advances in machine learning to tune search-based acquisition function optimizers on-the-fly to improve its accuracy in selecting candidate structures for evaluation.

There is also work on solving BO over discrete spaces by reduction to continuous BO \cite{reduction_continuous}. The key idea is to employ an encoder-decoder architecture to learn continuous representation from data and perform BO in this latent space. The main drawback of this method is that it generates a large fraction of invalid structures. This approach also requires a large database of ``relevant'' structures, for training an auto-encoder, which will not be available in many applications, where {\em small data} is the norm.

The challenge of optimizing acquisition function for continuous input spaces was  tackled in previous work \cite{max_acq_Neurips}. Since this approach relies on gradients for optimizing the acquisition function, it is specific to continuous  spaces and cannot be generalized to the challenging case of discrete spaces.
A tangential line of work \cite{meta_learning_acq_1,meta_learning_acq_2} exploiting the idea of learning acquisition function strategies across multiple tasks has also been explored in the context of transfer learning for black-box optimization in the BO framework. However, our problem setting is very different as we focus on learning within a single task when we have not yet solved the task.

\section{Learning to Search Framework}

In this section, we first motivate learning-to-search (L2S) methods for solving acquisition function optimization (AFO) problems. 
Subsequently, we describe our proposed learning to search framework, L2S-DISCO, and  provide a concrete instantiation for local-search based AFO.

\subsection{Motivation}

\vspace{1.0ex}

\noindent {\bf Search-based AFO solvers.} In a search-based optimizer, the overall problem-solving can be modeled as a computational search process defined in terms of an appropriate {\em search space} over candidate solutions, {\em search procedure} to uncover solutions, and {\em search control knowledge} to guide the search. 
For example, a solver, based on local search with multiple restarts, may use control knowledge that biases the restart distribution. Similarly, a solver, based on branch-and-bound search, may use control knowledge corresponding to policies for node expansion and pruning based on the current state of the solver. An important aspect of search-based optimization is that we can potentially improve the search control knowledge 
during a search by feeding the information about the search progress to machine learning techniques. 

{\em Relation between AFO problems.} We now give the intuition for why it may be useful to learn control knowledge across the sequence of AFO problems encountered during BO. Recall that the change in acquisition function $\mathcal{AF}(\mathcal{M},x)$ from iteration $i$ to $i+1$ is due to {\em only one} new training example $(x_i,y_i)$, where $x_i$ is the selected structure in iteration $i$ and $y_i$ is its function evaluation. Intuitively, even if the acquisition function scores of candidate structures in $\mathcal{X}$ are changing, the search control knowledge can still guide the search towards promising structures and only require small modifications to account for the slight change in the AFO problem from previous BO iteration. This motivates using machine learning to adapt the knowledge in a way that generalizes from prior iterations of AFO to future AFO iterations. 

\vspace{1.0ex}

\noindent {\bf Empirical evidence for the utility of learning.} We now provide some empirical evidence on real-world problems to show how machine learning can be potentially useful to improve the accuracy of solving AFO problems. We consider local search with multiple restarts as the AFO solver. In this case, the AFO solver takes as input the objective function $\mathcal{AF}(\mathcal{M},x)$ and 
restarting strategy, and returns the local optima $\hat{x} \in \mathcal{X}$ with associated acquisition function value $\mathcal{AF}(\mathcal{M},\hat{x})$. The accuracy of local search based AFO solver critically depends on the restart strategy. We performed local search based AF optimization using three different restart strategies on optimization problems with binary discrete variables: 1. Completely random ({\em random}); 2. Assigning the first four discrete variables as zero and remaining randomly ({\em first-four-zero}); and 3. Assigning the last four discrete variables as one and remaining randomly ({\em last-four-one}). In figure~\ref{fig:empirical_evidence}, we show the results of solving a single AFO problem using these three different restart strategies over 100 runs. We plot the distribution of $\mathcal{AF}(\mathcal{M},\hat{x})$ for these strategies. We can see that different restart strategies give varied solutions (empirically). This observation can be leveraged to learn a search heuristic to select promising starting states for local search using the training data from local search trajectories. 

\subsection{L2S-DISCO and Key Elements}

L2S-DISCO integrates machine learning techniques and combinatorial search in a principled manner for accurately solving AFO problems to select combinatorial structures for evaluation. This framework allows us to employ surrogate statistical models of arbitrary complexity and can work with any acquisition function. The key insight behind L2S-DISCO is to directly tune the search via learning during the optimization process to select the next structure for evaluation. The search-based perspective has several advantages: 1) High flexibility in defining search spaces over structures; 2) Easily handles domain constraints that determine which structures are ``valid''. For example, when designing an optimized network on the chip to facilitate data transfer between multiple cores, we need to make sure that there is a viable path between any pair of cores; 3) Allows to incorporate prior knowledge in the form of heuristic rules to explore promising regions of the search space; and 4) Provides additional points for learning within the search framework to improve the effectiveness of search in uncovering better structures. 

\vspace{1.0ex}
\begin{figure}[t!]
\centering
\includegraphics[width=\columnwidth]{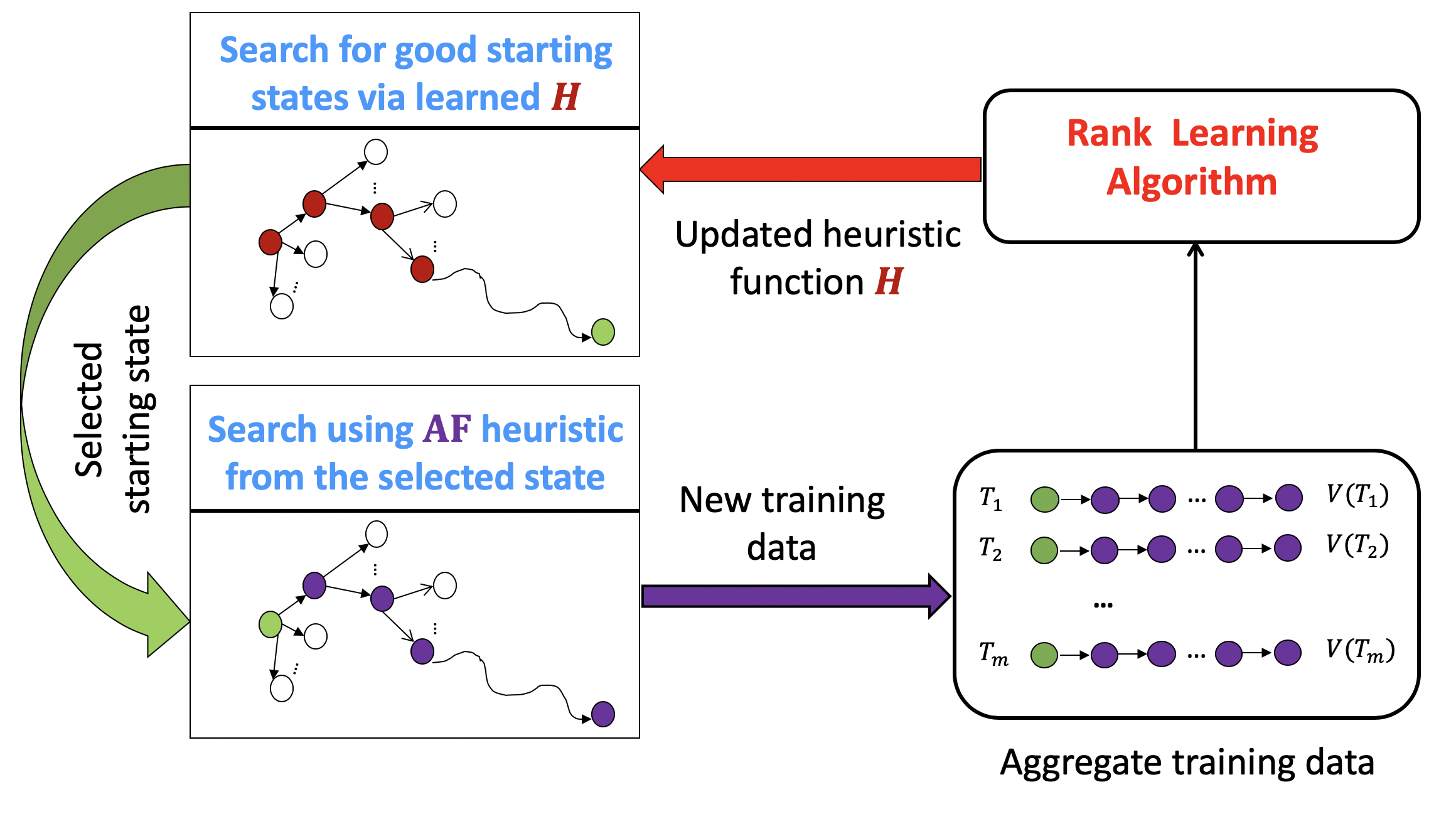}
\caption{High-level overview of L2S-DISCO instantiation for local search. It repeatedly performs three steps. First, run local search from a random state guided by current heuristic $\mathcal{H}$ to select a good starting state. Second, run local search from this selected starting state guided by acquisition function ($\mathcal{AF}$). Third, use new training data in the form of local search trajectory $T$ and acquisition function value of the local optima $V(T)$ to update the heuristic $\mathcal{H}$ via rank learning.}
\label{fig:main_fig}
\end{figure}
\noindent {\bf Overview of L2S-DISCO.} We build a surrogate model $\mathcal{M}$ using a small number of experiments and their outcomes to guide our search process to select the sequence of combinatorial structures to perform experiments. L2S-DISCO is parameterized by a search space $\mathcal{S}$ over structures, a learned function $\mathcal{AF}(\mathcal{M},x \in \mathcal{X})$ to score the utility of structures for evaluation, a search strategy $\mathcal{A}$ (e.g., local search), and a learned search control knowledge $\mathcal{H}$ to guide the search towards high-scoring structures. In each BO iteration, we perform the following two steps repeatedly until the maximum time-bound is exceeded or a termination criteria is met. {\bf Step 1:} Execute search strategy $\mathcal{A}$ guided by the current search control knowledge to uncover promising structures. {\bf Step  2:} Update the parameters of search control knowledge $\mathcal{H}$ using the online training data generated from the recent search experience. Fig~\ref{fig:main_fig} illustrates the instantiation of L2S-DISCO for local search. Each structure $x \in \mathcal{X}$ uncovered during the entire search is scored according to $\mathcal{AF}(\mathcal{M},x)$ and we select the highest scoring structure $x_{next}$ for function evaluation. We perform experiment using the selected structure $x_{next}$ and observe the outcome $\mathcal{F}(x_{next})$. The statistical model $\mathcal{M}$ is updated using the new training example $(x_{next}, \mathcal{F}(x_{next}))$. We repeat the next iteration of BO via L2S-DISCO initialized with the current search control knowledge. 

\vspace{1.0ex}

\noindent {\bf Key Elements.} There are two key elements in L2S-DISCO that need to be specified to instantiate it for a given search procedure. {\bf 1)} The form of training data to learn search control knowledge $\mathcal{H}$; and {\bf 2)} The learning formulation and associate learning algorithm to update the parameters of search control knowledge $\mathcal{H}$ using online training data. These elements vary for different search procedures and forms of search control knowledge. We provide a high-level example to illustrate these elements for branch-and-bound search. 

Branch-and-bound search is a widely used search procedure to solve combinatorial optimization  problems. It employs a search space over partial structures, where each state corresponds to partial assignment of variables. The states with complete assignment for all variables are referred as terminals. Variable selection strategy for successive assignment is one of the main components of branch-and-bound search. Therefore, 
$\mathcal{H}$ corresponds to the policy that selects the variable on which to branch on for the next assignment. In this case, the training data is generated by the trajectories obtained by a strong branching (SB) strategy \cite{learning_to_branch} which exhaustively tests each variable for assignment. A learning-to-rank formulation is natural for inducing the variable selection policy, since the reference strategy (SB) effectively ranks variables at a node by a score, and picks the highest-scoring variable, i.e., the score itself is not important. 

Below we provide a concrete instantiation of L2S-DISCO for local search based acquistion function optimization that will be employed for our empirical evaluation.

\begin{algorithm}[b!]
\caption{L2S-DISCO for local search}
\label{alg:local-search}
\footnotesize
\textbf{Input}: $\mathcal{X}$= space of combinatorial structures, 
$\mathcal{AF}(\mathcal{M}, x)$= acquisition function,
$\mathcal{H}(\theta, x)$= search heuristic from previous BO iteration,
$\textsc{RankLearn}$= rank learner \\
\textbf{Output}: $\hat{x}_{next}$, the selected structure for function evaluation
\label{alg:L2S-LS}
\begin{algorithmic}[1]
\STATE Initialization: $\mathcal{T} \leftarrow \emptyset$ (training data of local search trajectories) and $\mathcal{S}_{start}\leftarrow \emptyset$ (set of starting states)
\REPEAT
\STATE Perform local search from a random state $x \in \mathcal{X}$ guided by heuristic $\mathcal{H}(\theta, x)$ to reach a local optima $x_{restart}$
\IF{$x_{restart} \in \mathcal{S}_{start}$}
	\STATE $x_{start} \leftarrow$ random structure from $\mathcal{X}$
\ELSE
	\STATE $x_{start} \leftarrow x_{restart}$
\ENDIF
\STATE Perform local search from $x_{start}$ guided by $\mathcal{AF}(\mathcal{M},x)$ 
\STATE Add the new search trajectory and $\mathcal{AF}(\mathcal{M},x_{end})$ to $\mathcal{T}$
\STATE Update heuristic $\mathcal{H}(\theta, x)$ via rank learner using $\mathcal{T}$
\STATE $\mathcal{S}_{start} \leftarrow \mathcal{S}_{start} \cup x_{start}$ 
\UNTIL{convergence or maximum iterations}
\STATE $\hat{x}_{next} \leftarrow$ best scoring structure as per $\mathcal{AF}(\mathcal{M},x)$ found during the entire search process 
\STATE \textbf{return} the selected structure for evaluation $\hat{x}_{next}$
\end{algorithmic}
\end{algorithm}

\subsection{Instantiation of L2S-DISCO for Local Search}

Recall that local search based AFO solver performs multiple runs of local search guided by the acquisition function $\mathcal{AF}(\mathcal{M},x)$ from different random starting states. The search space is defined over complete structures, where each state corresponds to a complete structure $x \in \mathcal{X}$. The successors of a state with structure $x$ referred as $\mathcal{N}(x)$, is the set of all structures $x' \in \mathcal{X}$ such that the hamming distance between $x$ and $x'$ is one. The effectiveness of local search depends critically on the quality of starting states. Therefore, we instantiate L2S-DISCO for local search and learn a search heuristic $\mathcal{H}(\theta, x)$ to select good starting states that will allow local search to uncover high-scoring structures from $\mathcal{X}$ according to $\mathcal{AF}(\mathcal{M},x)$. 

To instantiate L2S-DISCO for local search, we need to specify the two key elements: 1) The training data for learning the heuristic $\mathcal{H}(\theta, x)$?; and 2) The learning formulation to induce $\mathcal{H}(\theta, x)$ 
from online training data.

\vspace{1.0ex}

\noindent {\bf 1) Training data.} The set of search trajectories $\mathcal{T}$ obtained by performing local search from different starting states and   acquisition function scores for local optima correspond to the training data. Each search trajectory $T \in \mathcal{T}$ consists of the sequence of states from the starting state $x_{start}$ to the local optima $x_{end}$. Suppose $V(T)$=$\mathcal{AF}(\mathcal{M},x_{end})$ represents the acquisition function score of the local optima for local search trajectory $T$.

\vspace{1.0ex}

\noindent {\bf 2) Rank learning formulation.} The role of the heuristic $\mathcal{H}(\theta, x)$ is to rank candidate starting states according to their utility in uncovering high-scoring structures from $\mathcal{X}$ via local search. Recall that if we perform local search guided by $\mathcal{AF}(\mathcal{M},x)$ from any state $x$ on a search trajectory $T \in \mathcal{T}$, we will reach the same local optima with acquisition function score $V(T)$. In other words, every state on the trajectory $T \in \mathcal{T}$ has the same utility. Therefore, we formulate the problem of learning the search heuristic as an instance of bipartite ranking \cite{BPR:COLT-2005}. Specifically, for every pair of search trajectories $T_1, T_2 \in \mathcal{T}$, if $V(T_1) > V(T_2)$, then we want to rank every state on the trajectory $T_1$ better than every state on the trajectory $T_2$. We will generate one ranking example for every pair of states $(x_1, x_2)$, where $x_1$ is a state on the trajectory $T_1$ and $x_2$ is a state on the trajectory $T_2$. The aggregate set of ranking examples are given to an off-the-shelf rank learner to induce $\mathcal{H}(\theta, x)$, where $\theta$ are the parameters of the ranking function. In our experiments, we employed RankNet \cite{ranknet} as the base rank learner. We leveraged existing code\footnote{\url{https://github.com/shiba24/learning2rank}} for our purpose.

\vspace{3.0ex}

\noindent {\bf L2S-DISCO for local search based optimization.} Figure~\ref{fig:main_fig} illustrates L2S-DISCO instantiation for local search based acquisition function optimization. At a high-level, each iteration of L2S-DISCO consists of two alternating local search runs. First, local search guided by heuristic $\mathcal{H}$ to select the starting state. Second, local search guided by $\mathcal{AF}$ from the selected starting state. After each local search run, we get a new local search trajectory, and the heuristic function $\mathcal{H}$ is updated to be consistent with this new search trajectory.

Algorithm~\ref{alg:local-search} shows the pseduo-code for learning based local search to solve AFO problems arising in BO iterations. It reuses the learned search heuristic from the previous BO iteration and updates it in an online manner using the new training data generated during AF optimization. In each iteration, we perform the following sequence of steps. First, we perform local search from a random state guided by the search heuristic $\mathcal{H}(\theta, x)$  until reaching the local optima $x_{restart}$ to select the next starting state. Second, if $x_{restart}$ was not explored as a starting state in previous local search iterations, we select $x_{restart}$ as the starting state to perform local search guided by $\mathcal{AF}(\mathcal{M},x)$ and add the local search trajectory to our training data. Third, we update the search heuristic $\mathcal{H}(\theta, x)$ using the newly added training example via rank learner. We repeat the above three steps until convergence or maximum iterations. This instantiation of L2S-DISCO is similar in spirit to the STAGE algorithm \cite{STAGE}. At the end, we return the best scoring structure uncovered during the search $\hat{x}_{next}$ for function evaluation.

\section{Experiments and Results}
\begin{figure*}[t]
\centering
\subfloat[Subfigure 1 list of figures text][Contamination domain  with no. of stages $d=25$  and $\lambda=10^{-4}$  over 250 iterations.]{
\includegraphics[width=0.45\textwidth]{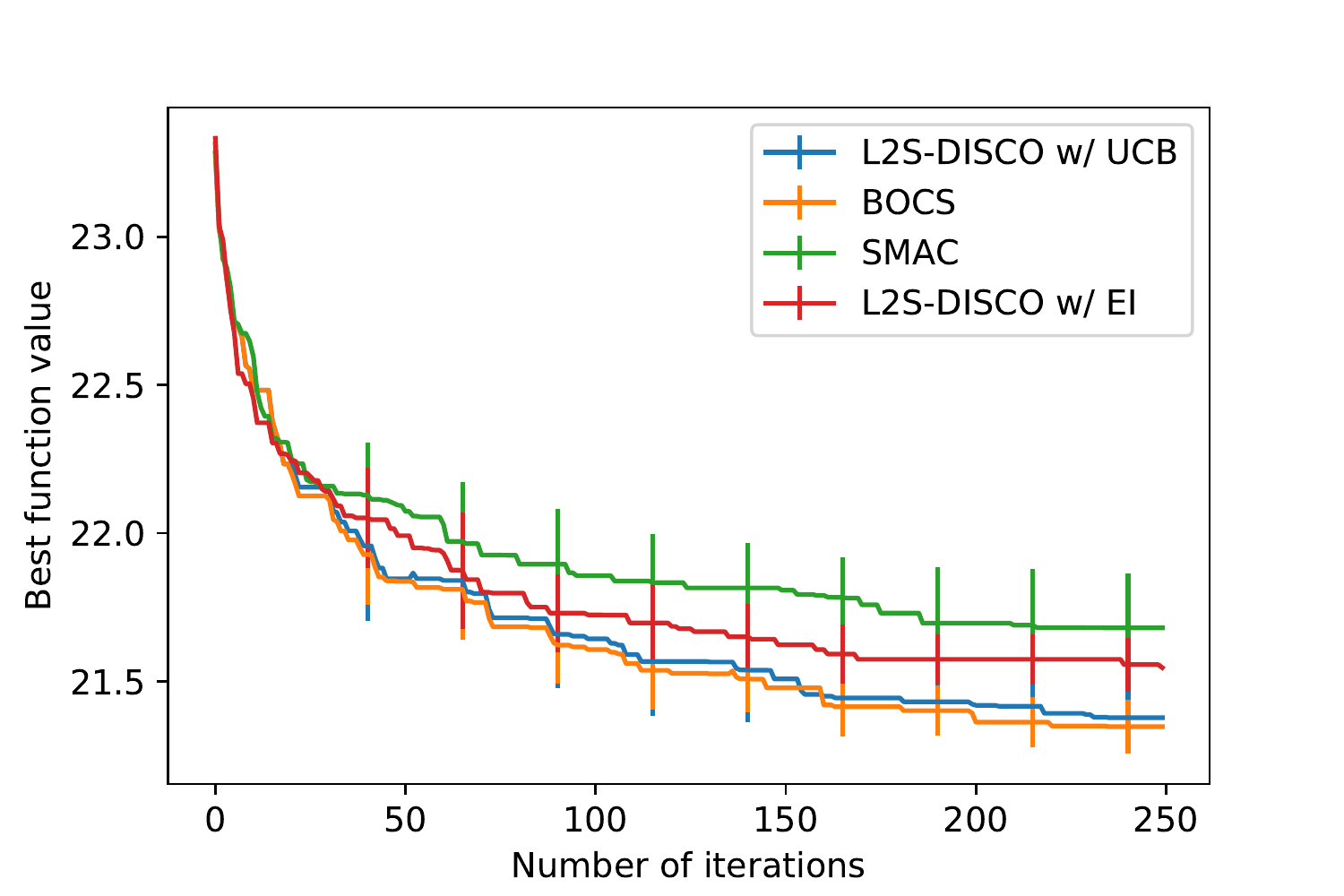}
\label{fig:contamination}}
\qquad
\subfloat[Subfigure 2 list of figures text][Ising domain with number of nodes $d=24$  and $\lambda=10^{-2}$  over 150 iterations.]{
\includegraphics[width=0.45\textwidth]{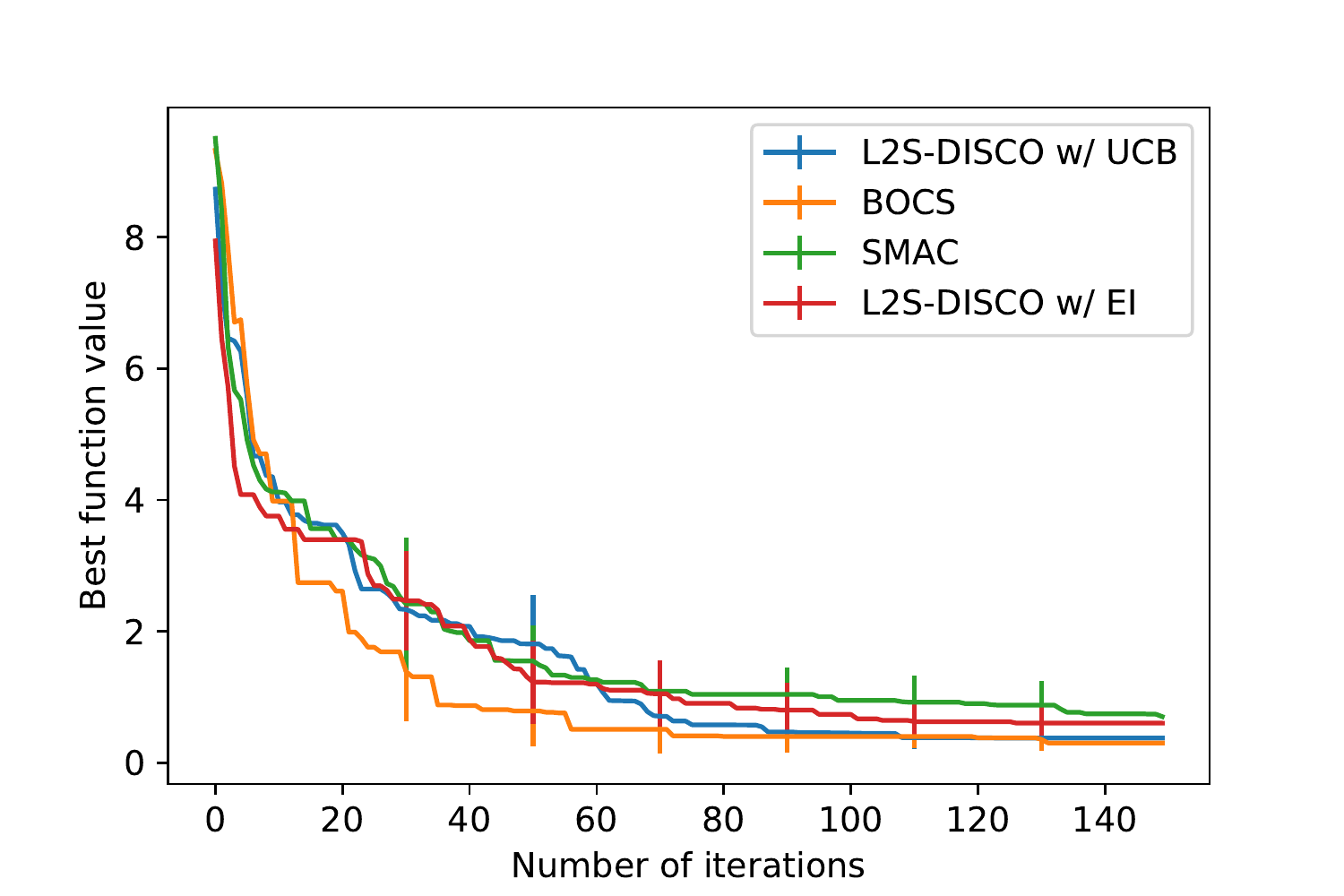}
\label{fig:ising}}
\caption{Results for contamination and ising  domain ({\bf minimization}).} 
\label{fig:cont_ising_fig}
\end{figure*}
In this section, we first describe our experimental setup and then discuss the results of L2S-DISCO and baseline methods.

\subsection{Experimental Setup}

\noindent {\bf Benchmark Domains.} We employ five diverse benchmark domains for our empirical evaluation.

\vspace{0.2ex}

{\bf 1. Contamination.} The problem considers a food supply with $d$ stages, where a binary \{0,1\} decision must be made at each stage to prevent the food from being contaminated with pathogenic micro-organisms \cite{contamination,BOCS}. Each prevention effort at stage $i$ can be made to decrease the contamination by a given random rate $\Gamma_i$ and incurring a cost $c_i$. The contamination spreads with a random rate $\Lambda_i$ if no prevention effort is taken. The overall goal is to ensure that the fraction of contaminated food at each stage $i$ does not exceed an upper limit $U_i$ with probability at least $1-\epsilon$ while minimizing the total cost of all prevention efforts. Following \cite{BOCS}, the lagrangian relaxation based problem formulation is: 
\begin{align*}
    \arg \min_x \sum_{i=1}^d \left[c_i x_i + \frac{\rho}{T} \sum_{k=1}^T 1_{\{Z_k > U_i\}} \right] + \lambda \|x\|_1
\end{align*}
where $\lambda$ is a regularization coefficient, $Z_i$ is the fraction of contaminated food at stage $i$, violation penalty coefficient $\rho$=$1$, and $T$=$100$.

\vspace{0.2ex}

{\bf 2. Sparsification of zero-field Ising models.} The distribution of a  zero field Ising model $p(z)$ for $z \in \{-1, 1\}^n$ is characterized by a symmetric interaction matrix $J^p$ whose support is represented by a graph $G^p = ([n], E^p)$ that satisfies $(i,j) \in E^p$ if and only if $J^p_{ij} \neq 0$ holds \cite{BOCS}. The overall goal 
is to find a close approximate distribution $q(z)$ while minimizing the number of edges in $E^q$. Therefore, the objective function in this case is a regularized KL-divergence between $p$ and $q$ as given below:
\begin{align*}
    D_{KL} (p || q_x) = \sum_{(i,j) \in E^p} (J_{ij}^p - J_{ij}^q) E_p[z_iz_j] + log (Z_q/Z_p)
\end{align*}
where $Z_q$ and $Z_p$ are partition functions corresponding to $p$ and $q$ respectively, and $x \in \{0,1\}^{E^q}$ is the decision variable representing whether each edge is present in $E^q$ or not.

\vspace{0.2ex}

{\bf 3. Low auto-correlation binary sequences (LABS).} The problem is to find a binary \{+1,-1\} sequence $S$ = $(s_1, s_2,\cdots, s_n)$ of given length $n$ that maximizes {\it merit factor} defined over a binary sequence as given below:
\begin{align*}
\text{Merit Factor(S)} &= \frac{n^2}{E(S)} \hspace{1mm}\\
\text{where} \hspace{1mm} E(S) &= \sum_{k=1}^{n-1} \left(\sum_{i=1}^{n-k} s_i s_{i+k}\right)^2
\end{align*}
The LABS problem has multiple applications in diverse scientific disciplines \cite{LABS}.

\vspace{0.2ex}

{\bf 4. Network optimization in multicore chips.} With Moore's law aging quickly, multicore architectures are considered very promising for parallel computing \cite{Arch2030}. A key challenge in multicore research is to reduce the performance bottleneck due to data movement. One promising solution  is to optimize the placement of communication links between cores to facilitate efficient data transfer. This optimization is typically guided by expensive simulators that mimics the real hardware. The network optimization problem is part of the rodinia benchmark \cite{rodinia-benchmark} and uses the gem5-GPU simulator \cite{power:gem5-gpu:cal:2014}. There are 12 cores whose placements are fixed and the goal is to place 17 links between them to optimize performance: {\em 66 binary variables}. There is one {\em constraint} to determine valid structures: existence of a viable path between any pair of cores. We report the performance improvement with respect to the provided baseline network. 

\vspace{0.2ex}

{\bf 5. Core placement optimization in multicore chips.} This is another multicore architecture optimization problem from rodinia benchmark \cite{rodinia-benchmark}. In this problem, we are given 64 cores of three types (8 CPUs, 40 GPUs, and 16 memory units) and they are connected by a mesh network (every core is connected to its four neighboring cores) to facilitate data transfer. The goal is to place the three types of cores to optimize performance: {\em 64 categorical variables} with each taking three candidate values. We need to make sure that the {\em cardinality constraints} in terms of the number of cores of each type are satisfied. We report the performance improvement w.r.t the provided baseline placement. 

\vspace{1.0ex}

\noindent {\bf Baseline Methods.} We compare the local search instantiation of L2S-DISCO with two state-of-the-art methods: SMAC \cite{SMAC:TR2010} and BOCS \cite{BOCS}. We employed open-source python implementations of both BOCS \footnote{\url{https://github.com/baptistar/BOCS}} and SMAC \footnote{\url{https://github.com/automl/SMAC3}}. Since SMAC implementation does not support handling domain constraints to search over valid structures\footnote{https://github.com/automl/SMAC3/issues/403}, we could not run SMAC for network optimization and core placement optimization benchmarks. Similarly, SDP based solver for BOCS cannot handle constraints, so we employed simulated annealing based solver available in the BOCS code for those two benchmarks. We initialize the surrogate of all the methods by evaluating 20 random structures. For L2S-DISCO. we employed random forest model with 20 trees (tried two standard settings of scikit-learn library, namely, 10 and 20 trees, and got similar results) and two different acquisition functions (EI and UCB). For UCB, we use the adaptive rate recommended by \cite{UCB:ICML2010} to set the exploration and exploitation trade-off parameter $\beta_i$ value depending on the iteration number $i$. We ran L2S-DISCO (Algorithm~\ref{alg:local-search}) for a maximum of 60 iterations. 

\vspace{1.0ex}

\noindent {\bf Evaluation Metric.} We use the best function value achieved after a given number of iterations 
as a metric to evaluate all methods: SMAC, BOCS, and L2S-DISCO. The method that uncovers high-performing structures with less number of function evaluations is considered better. LABS is a maximization problem, but the remaining four benchmarks require the objective to be minimized. We use the total number of iterations similar to BOCS \cite{BOCS}.

\subsection {Results and Discussion}

We discuss the results of L2S-DISCO and baseline methods on the five benchmarks below. All the reported results are averaged over 10 random runs (except for BOCS in cores placement optimization due to its poor scalability).

\subsubsection{Contamination and Ising.} Figure \ref{fig:cont_ising_fig} shows the comparison of L2S-DISCO with SMAC and BOCS baselines. 
We make the following observations. 1) Both L2S-DISCO variants that use EI and UCB acquisition functions perform better than SMAC. 2) L2S-DISCO with UCB performs better than the variant with EI. We observed a similar trend for the remaining three benchmarks also. Therefore, to avoid clutter, we only show the results of L2S-DISCO with UCB for the remaining benchmarks. 3) Results of L2S-DISCO are comparable to BOCS on the contamination problem. However, BOCS has a better anytime profile for ising domain. L2S-DISCO eventually matches the performance of BOCS after 90 iterations. The main reason BOCS performs 
slightly better in these two domains is that they exactly match the modeling assumptions of BOCS, which allows the use of SDP based solver to select structures for evaluation. Below we will show how the performance of BOCS degrades when the assumptions are not met, whereas L2S-DISCO peforms robustly across optimization problems of varying complexity.

\begin{figure}[h!]
\centering
\includegraphics[width=\columnwidth]{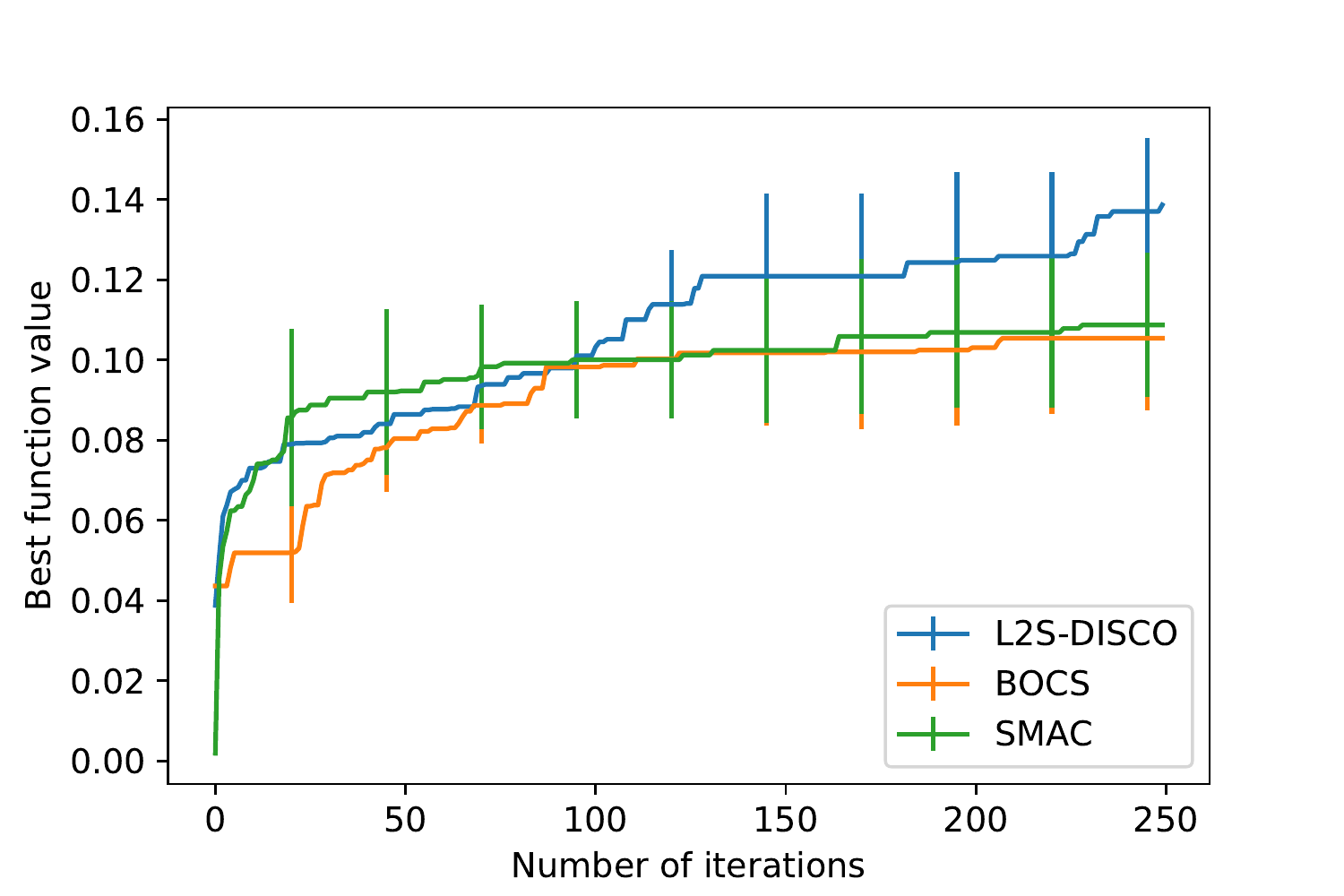}
\caption{Results for LABS  domain ({\bf maximization}) with input sequence length $n$=30 over 250 iterations.}
\label{fig:labs_fig}
\end{figure}
\subsubsection {LABS.} Figure \ref{fig:labs_fig} shows the comparison of L2S-DISCO with SMAC and BOCS baselines. 
We can see that L2S-DISCO clearly outperforms both BOCS and SMAC on this domain. BOCS has the advantage of SDP based solver, but its statistical model that accounts for only pair-wise interactions is limiting to account for the complexity in this problem. SMAC and L2S-DISCO both employ random forest model, but L2S-DISCO does better in terms of acquisition function optimization by integrating learning with search.

\begin{figure}[h!]
\centering
\includegraphics[width=\columnwidth]{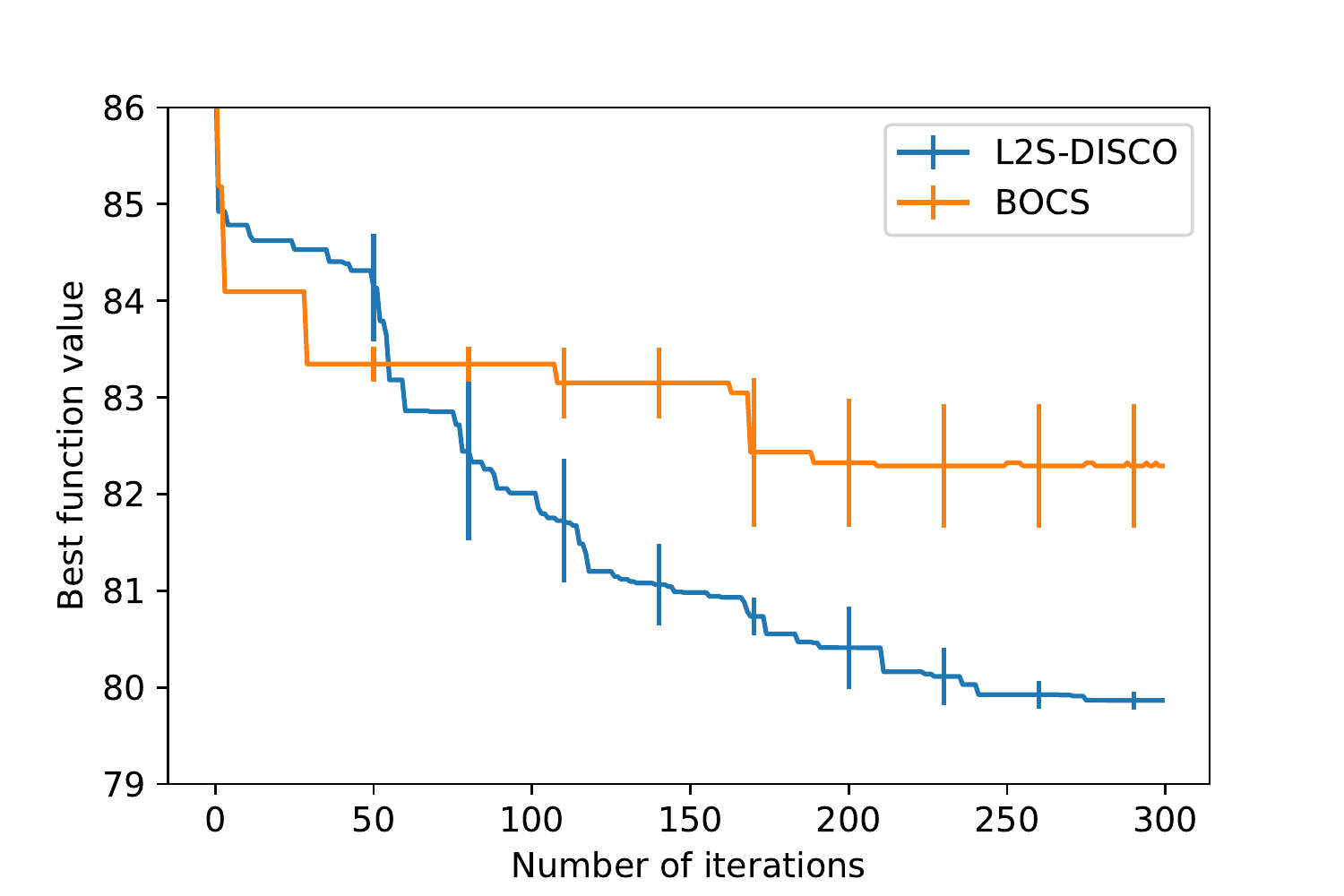}
\caption{Results for network optimization in multicore chips ({\bf minimization}) over 300 iterations.}
\label{fig:noc_links_fig}
\end{figure}

\subsubsection {Network optimization in multicore chips.} As mentioned earlier, we could not run SMAC for this problem as SMAC library does not allow to incorporate complex domain constraints. The SDP solver of BOCS is also not applicable due to complex constraints. Hence, we employ simulated annealing based solver for acquisition function optimization. Figure \ref{fig:noc_links_fig} shows the comparison of L2S-DISCO with BOCS baseline. We can see that L2S-DISCO performs significantly better than BOCS in this domain.  BOCS seems to get stuck for long periods, whereas L2S-DISCO shows consistent improvement in uncovering high-performing structures. This behavior of BOCS can be partly attributed to the limitations of both surrogate model and acquisition function optimizer.

\subsubsection {Core placement optimization in multicore chips.} Figure \ref{fig:noc_links_fig} shows the comparison of L2S-DISCO with BOCS. L2S-DISCO significantly outperforms BOCS on this benchmark also. Additionally, BOCS scales poorly on this domain, where the discrete variables are non-binary. Recall that BOCS model was developed for binary variables and authors suggested the use of one-hot encoding to handle categorical variables. However, this transformation excessively increases the no. of dimensions. For example, we have 64 dimensions for L2S-DISCO, but it grows to 192 for BOCS due to one-hot encoding and makes its execution extremely slow. BOCS took one hour per single BO iteration on a machine with Intel Xeon(R) 2.5Ghz CPU and 96 GB memory. This is the main reason we could only perform one run of BOCS.
\begin{figure}[h!]
\centering
\includegraphics[width=\columnwidth]{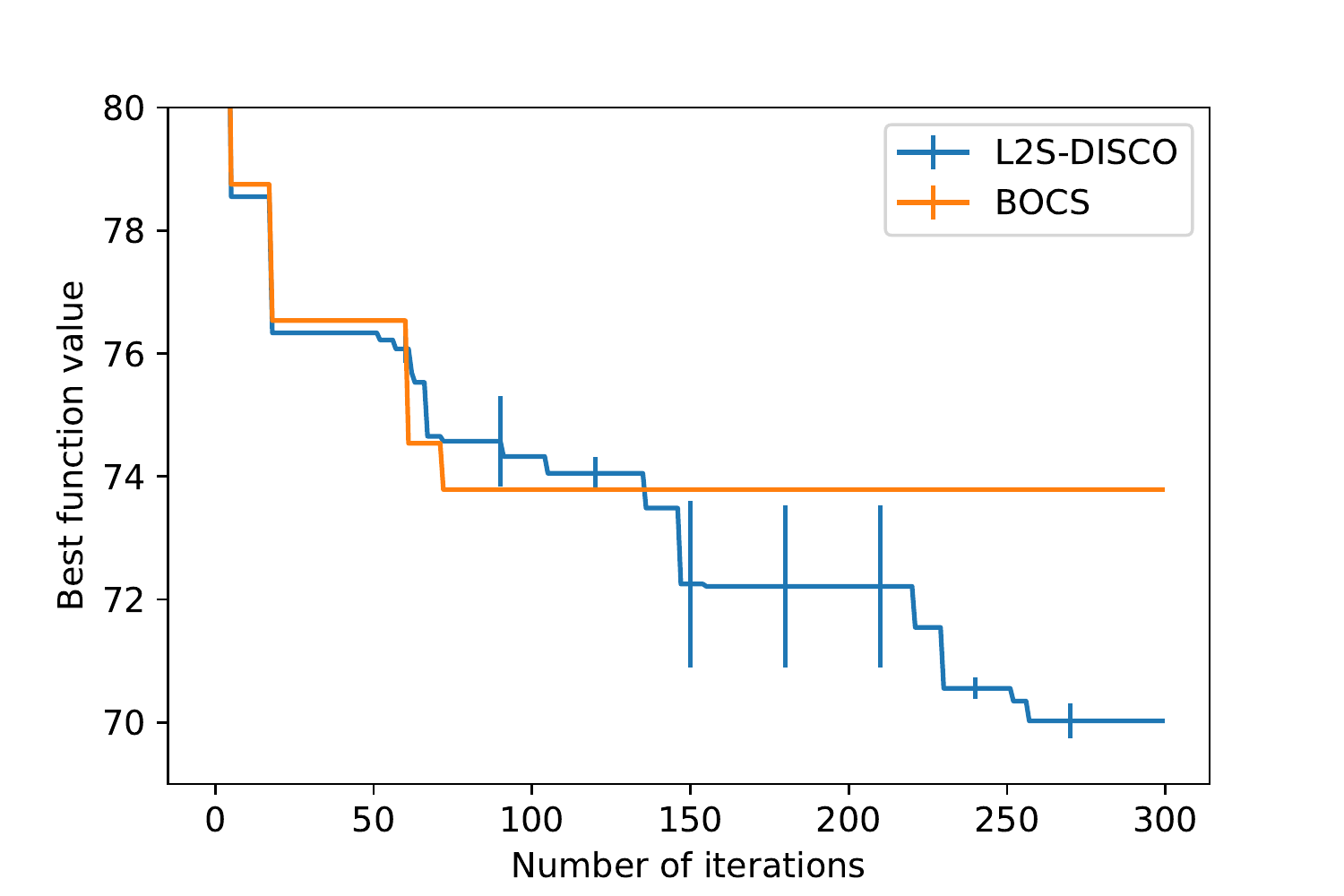}
\caption{Results for core placement optimization in multicore chips ({\bf minimization}) over 300 iterations.}
\label{fig:cores_fig}
\end{figure}

\section{Summary and Future Work}

We introduced the L2S-DISCO framework that integrates machine learning with search-based optimization for optimizing expensive black-box functions over discrete spaces. We showed that instantiation of L2S-DISCO for local search based optimization yields significantly better performance than state-of-the-art methods on complex optimization problems. Future work includes 
studying instantiations of L2S-DISCO for other search procedures to further improve the performance, and applying L2-DISCO on important real-world applications by leveraging domain knowledge.

\vspace{1.0ex}

\footnotesize
\noindent {\bf Acknowledgements.} We thank the anonymous reviewers for their feedback. This research is supported by National Science Foundation grants IIS-1845922, OAC-1910213, and IIS-1619433.

\bibliographystyle{aaai}
\bibliography{references}
\end{document}